\documentclass{esannV2}
\usepackage[dvips]{graphicx}
\usepackage[utf8]{inputenc}
\usepackage{amssymb,amsmath,array,bm}
\usepackage{graphicx,tabularx,booktabs}
\usepackage{mathabx}
\usepackage{algorithm}
\usepackage{algorithmicx}
\usepackage{algpseudocode}
\usepackage{color}
\usepackage{url}
\usepackage{enumitem} 
\usepackage{multirow, multicol}
\usepackage{wrapfig}
\usepackage{hyperref}

\newcommand{\norm}[1]{\left\lVert#1\right\rVert}
\newcommand*{\placeholderfunc}{\makebox[1ex]{\textbf{$\cdot$}}}%
\begin{document}

\newcommand{\algoname}{TBD}

\title{Sparse Uncertainty-Informed Sampling \\ from Federated Streaming Data}

\author{Manuel R\"oder\textsuperscript{1,2,3} and Frank-Michael Schleif\textsuperscript{1}
%
\thanks{MR is supported through the Bavarian HighTech Agenda, specifically by the Würzburg Center for Artificial Intelligence and Robotics (CAIRO) and the ProPere THWS scholarship.} 
%
\vspace{.3cm}\\
%
1 - Technical UAS W\"urzburg-Schweinfurt, Fac. of Computer Science,  W\"urzburg, DE \\
2 - Bielefeld University, Faculty of Technology, Bielefeld, DE \\
3 - Center for Artificial Intelligence and Robotics, Würzburg, DE
}


\maketitle

\begin{abstract}
We present a numerically robust, computationally efficient approach for non-I.I.D. data stream sampling in federated client systems, where resources are limited and labeled data for local model adaptation is sparse and expensive. 
The proposed method identifies relevant stream observations to optimize the underlying client model, given a local labeling budget, and performs instantaneous labeling decisions without relying on any memory buffering strategies. 
Our experiments show enhanced training batch diversity and an improved numerical robustness of the proposal compared to existing strategies over large-scale data streams, making our approach an effective and convenient solution in FL environments. 
\end{abstract}

\section{Introduction} 

The rapid evolution of digital technologies and the exponential growth of data, coupled with the society's increased demand for data protection, have underscored the significant importance of \emph{Federated Learning} (\textbf{FL})~\cite{mcmahan2017} in the modern AI era.
This innovative approach to machine learning embodies the shift towards decentralised data processing that preserves privacy and enables collaborative learning without direct data exchange.
The relevance of FL extends across a wide range of sectors, particularly where data protection, real-time analytics and security are crucial, such as healthcare, finance, automotive and manufacturing~\cite{rieke2020, long2020, zhang2021}. 
The necessary extension of FL to \emph{streaming data}~\cite{heusinger2023} in this context poses new challenges due to its real-time nature and source variability, demanding instant decision making and model adaptation in high-volume data processing pipelines that operate under resource contraints~\cite{marfoq2023}.
%
Given the voluminous nature of data streams and the high costs of data labeling, it is impractical and inefficient to process every data point for model training.
Selective sampling plays a critical role in addressing this issue by identifying and selecting the most beneficial data points for model updates, thereby optimising the learning process.
In general, one key challenge is to develop mechanisms that can perform this selection accurately and rationally in real-time, while remaining numerically stable despite iterating over thousands of data points.
%
Efficiently determining which data points warrant expert labeling becomes a pivotal aspect of the learning process, particularly in scenarios where labeled data is scarce or costly to acquire. The ability to strategically focus expert intervention on the most impactful data points is extensively researched in the field of \emph{Active Learning}~\cite{settles2009} and can significantly accelerate the learning process and enhance the overall effectiveness of the FL system. 


In this study, we introduce an advanced method that empowers resource-limited FL clients to select the most valuable samples to refine their neural network models from streaming data in real-time.
This approach prioritizes resource efficiency and numerical stability.
To achieve this, we employ a query strategy derived from \emph{Volume Sampling for Streaming Active Learning}~\cite{saran2023}, which makes labeling choices based on the penultimate layer representations, within a conformal FL framework. We modify this strategy for the FL streaming context by substituting the iterative Woodbury Identity update mechanism~\cite{max1950} -- which adjusts sampling probabilities -- with low-rank updates to the lower Cholesky triangular matrix~\cite{seeger2004}. This modification simplifies the decision process and ensures numerical stability across a vast array of streamed data points.

\section{Methodology}\label{sec:method}

\subsection{Setup and Prerequisites}\label{subsec:setup}
\begin{figure*}
    \centering
    \includegraphics[width=0.9\textwidth]{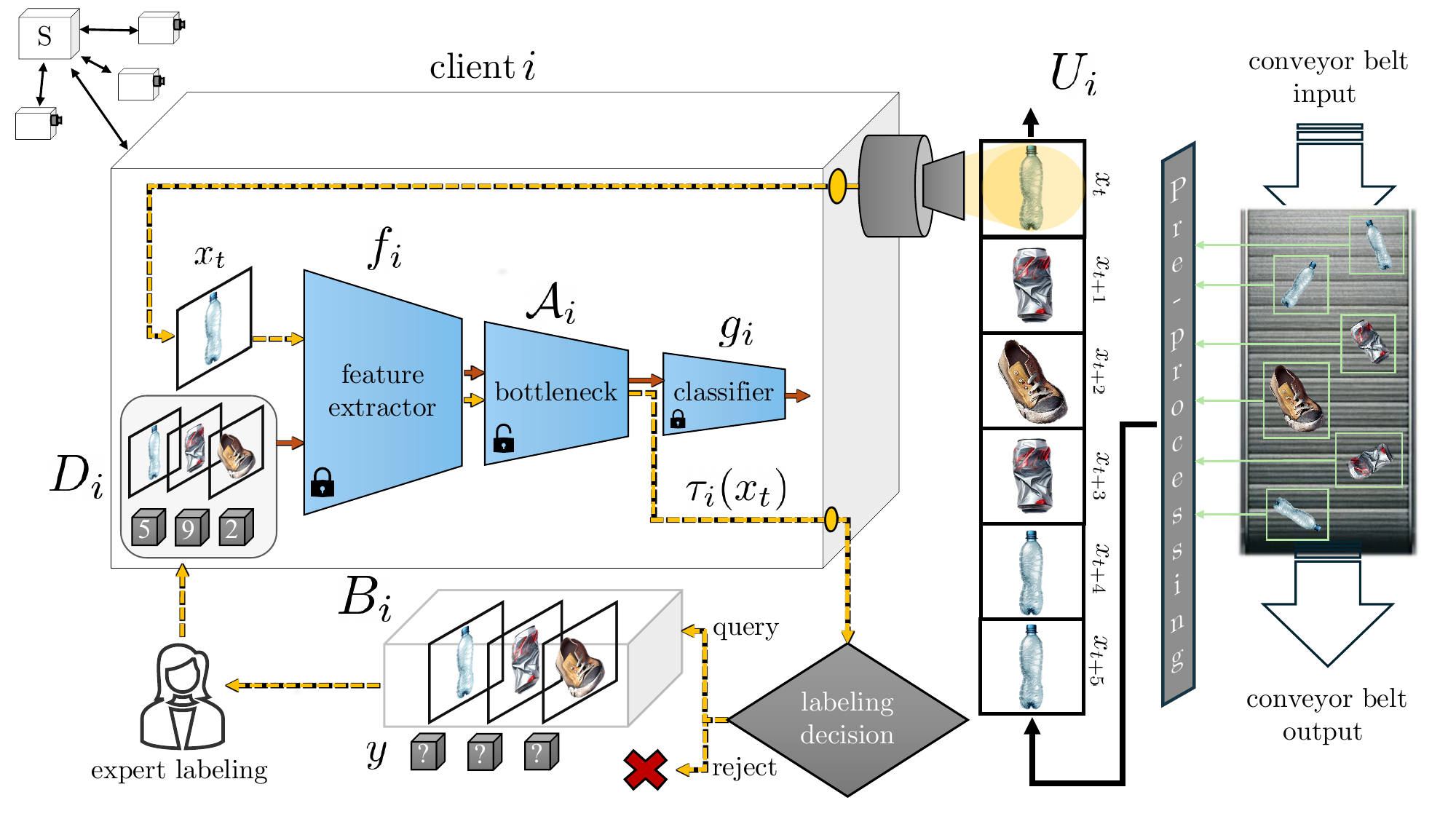}\vspace{-.55cm}
    \caption{Sparse sample selection from federated streaming data, examplified by conveyor belt object scanning on client $i$: the orange, dashed data flow illustrates the decision-making algorithm; the red data flow depicts the fine-tuning pipeline.}
    \label{fig:fig_1}
\end{figure*}

We consider a FL environment characterized by a central, resource-rich server $S$ tasked with network orchestration and multiple resource-limited clients indexed by $i$, where $i=1, \ldots I$ as outlined in Fig.~\ref{fig:fig_1}.
Clients operate without long-term storage, demanding immediate labeling decisions - a unique challenge. The client model features a core sequence of three components, including a pre-trained feature extractor
$f(\placeholderfunc)$, a bottleneck module $\mathcal{A(\psi)}$, and a linear classification head $g(\placeholderfunc)$.
Given an input sample $x$ observed on client $i$, the overall decision function is then expressed as $\mathcal{F}_i(x) = g(\mathcal{A}_{\psi_i}(f(x)))$. 
This configuration is aligned with (but not limited to) the principles outlined in FedAcross~\cite{roeder2024}, a scalable FL framework designed to handle client adaptation tasks with scarcely labeled target data.
Furthermore, the client $i$ inspects sample $x_t$, seen only once, from the \emph{non-I.I.D.} data stream $U_i$ at time $t$, in which the occurrence of objects is not evenly distributed, using an arbitrary sensing device.
To address the high labeling costs and limited resources for a single round of FL on a client edge device, we introduce a labeling budget $k$ that limits the set of selected samples $B_i$, such that $|B_i| \leq k$ applies.
After exhausting the labeling budget $k$ in the current FL round, we obtain labels for all samples in $B_i$ from an oracle and refine the client model by optimizing the cross-entropy loss between $\mathcal{F}_i(\placeholderfunc)$'s output and the acquired labels.

\noindent
\textbf{Main contributions:} We provide an efficient sampling strategy that performs an instant decision about current stream sample $x_t$ being included in $B_i$ or not whilst \textit{a}) sticking to the labeling budget $k$ to address local resource constraints, \textit{b}) enforcing batch diversity over all samples within $B_i$ to maximize training efficiency of the underlying client model and \textit{c}) keeping the algorithmic response stable and immediate over a large number of iterative processing steps to enable unsupervised stream observation over extended time periods.

\subsection{Algorithm}\label{subsec:algo}
The proposed algorithm for deciding whether or not to request a label for sample $x_t$ on client $i$ performs an uncertainty measure on its representation in the embedding space of the penultimate layer of the underlying model.
Therefore, the employed embedding function is defined over the client-side feature extractor $f_i$ and the bottleneck layer $\mathcal{A}_i$ using
$
\tau_i(x_t) = \mathcal{A}_i(f_i(x_t)), \; \tau_i(x_t) \in \mathbb{R}^{d},
$
with $d$ defining the output dimension of the penultimate network layer. 
Our method follows the volume sampling technique proposed in~\cite{saran2023} by choosing a sample $x_t$ for labeling with a probability $p$, proportional to its gradient's determinantal contribution.
Assuming that the underlying client data distribution is non-stationary, the probability $p_t$ is then computed to adhere for the streaming setting with
\begin{equation} \label{eq_prob}
    p_t = \frac{q_t \cdot \tau_i(x_t)^T \hat{\mathbf{\Sigma}}_t^{-1} \tau_i(x_t)}{\text{tr}\left( \frac{1}{t} \hat{\mathbf{\Sigma}}_t^{-1} \sum_{i=1}^{t} \tau_i(x_t)\tau_i(x_t)^T \right)},
\end{equation}
where $\hat{\mathbf{\Sigma}}_t^{-1}$ denotes the inverse of the tracking covariance matrix $\mathbf{A}_t \in \mathbb{R}^{d \times d}$ over samples already included in $B_i$ and $q_t$ refers to the adaptive labeling frequency for the current stream observation defined as $q_t = (k - |B_i|) / (|U_i| - t)$. 

To effectively update and inverse $\mathbf{A}$ for each new sample in the limited-resource regime of a federated client, we 
employ a \emph{Cholesky decomposition-based low-rank update mechanism}~\cite{seeger2004}.
We argue that alternative solutions driven by Sherman-Morrison-Woodbury formula to update the tracking covariance matrix after positive rank-1 updates suffer from numerical instabilities, rendering it impractical for large-scale stream sampling in industrial applications.

Hence, for $t=0$, the algorithm initializes the lower triangular matrix $\mathbf{L}_t$ by calculating the Cholesky decomposition of the (symmetric, positive-definite) tracking covariance matrix $\mathbf{A}_t$ with $\mathbf{A}_t = \mathbf{L}_t {\mathbf{L}_t}^T$.
After determination by Equation~\eqref{eq_prob} that the currently observed sample $x_t$ is eligible to be included in $B_i$, the updated inverse of the tracking covariance matrix $\hat{\mathbf{\Sigma}}_{t+1}^{-1}$ is obtained in two steps: First, the algorithm performs a positive rank-1 update by adding the sample's embedded representation $\tau_i(x_t)$ to the Cholesky factor $\mathbf{L}_t$ with
\begin{equation} \label{eq_rank_1}
    \mathbf{L}_{t + 1} = \mathbf{L}_t + \tau_i(x_t) (\tau_i(x_t))^T.
\end{equation}\vspace{-.2cm}
Noting that Equation~\eqref{eq_rank_1} can be re-written to
\begin{equation} \label{eq_givens_1}
    \mathbf{L}_{t + 1} = 
    \left[
        \begin{array}{c}
        {\mathbf{L}_t}^T \\
        {\tau_i(x_t)}^T
        \end{array}
    \right]^T
    \left[
        \begin{array}{c}
        {\mathbf{L}_t}^T \\
        {\tau_i(x_t)}^T
        \end{array}
    \right],\qquad
     \mathbf{Q}^T
     \left[
         \begin{array}{c}
         {\mathbf{L}_t}^T \\
         {\tau_i(x_t)}^T
       \end{array}
     \right]
     =
     \left[
         \begin{array}{c}
         {\mathbf{L}_{t+1}} \\
         0
         \end{array}
     \right]
\end{equation}
this problem can be efficiently computed as a product of \emph{Givens} rotations $\mathbf{Q} = \mathbf{Q}_1 \cdots \mathbf{Q}_n$, 
see~\cite{golub2013}. 
Taking advantage of the retained lower triangular structure of $\mathbf{L}_{t+1}$, the second step to refactor $\hat{\mathbf{\Sigma}}_{t+1}^{-1}$ involves calculating the inverse of the updated Cholesky factor using
$
\hat{\mathbf{\Sigma}}_{t+1}^{-1} = (\mathbf{L}_{t+1} (\mathbf{L}_{t+1})^T)^{-1},
$
which can be solved efficiently by employing an algorithm based on repeated back-substitions~\cite{seeger2004}.
Subsequently, after having observed all elements of $U_i$, domain expert labeling is queried for selected samples in $B_i$ and the labeled batch is added to the training data set $D_i$ for model adaptation on client $i$. 
A pseudocode is given in Alg.~\ref{alg:fed_sampling}.
\begin{algorithm}
    \caption{\small Federated Stream Sampling\label{alg:fed_sampling}}
    \begin{algorithmic}[1]\small
        \Require Embedding function $\tau_i(x) = \mathcal{A}_i(f_i(x))$ on client $i$, unlabeled stream of samples $U_i$, adaptive sampling rate $q$, labeling budget $k$
        \State Initialize $t = 1$, $\hat{\mathbf{\Sigma}}_0^{-1} = \lambda^{-1} \mathbf{I}_d$ \Comment{regularized by $\lambda$ for stability}
        \State Initialize $\mathbf{A}_0 = \mathbf{0}_{d,d}$ \Comment{covariance over all data}
        \State Initialize $B_i = \{\}$ with $|B_i| \leq k$ upon additions \Comment{limit selected samples}
        \State Initialize $\mathbf{L}_0$ = \texttt{\textsc{chol}}($\mathbf{A}_0$) \Comment{Cholesky lower triangular factorization}
        \For{$x_t \in U_i$}
            \State $\mathbf{A}_t \leftarrow \frac{t-1}{t} \mathbf{A}_{t-1} + \frac{1}{t} \tau_i(x_t) \tau_i(x_t)^T$ \Comment{covariance matrix update}
            \State $p_t = q \cdot \tau_i(x_t)^T \hat{\mathbf{\Sigma}}_{t}^{-1} \tau_i(x_t) \text{tr}(\hat{\mathbf{\Sigma}}_{t}^{-1} \mathbf{A}_t)^{-1}$ \Comment{sampling prob. according to Eq.~\eqref{eq_prob}}
            \State \textbf{with probability} $\min(p_t, 1)$:
            \State \hspace*{\algorithmicindent} $B_i \leftarrow B_i \cup \{x_t\}$
            \State \hspace*{\algorithmicindent} $\mathbf{L}_{t+1} \leftarrow \texttt{\textsc{cholupdate\_low\_rank}}(\mathbf{L}_t, \tau_i(x_t))$ \Comment{update according to Eq.~\eqref{eq_givens_1}}
            \State \hspace*{\algorithmicindent} $\hat{\mathbf{\Sigma}}_{t+1}^{-1} \leftarrow (\mathbf{L}_{t+1} (\mathbf{L}_{t+1})^T)^{-1}$
            \State \textbf{else}:
            \State \hspace*{\algorithmicindent} $\hat{\mathbf{\Sigma}}_{t+1}^{-1} \leftarrow \hat{\mathbf{\Sigma}}_{t}^{-1}$
        \EndFor
        \State \texttt{\textsc{query\_labels}}($B_i$) \Comment{request domain expert labeling}
    \State \Return labeled batch $B_i$ for client $i$ model adaptation
    \end{algorithmic}
\end{algorithm}

\section{Experiments}\label{sec_exp}
%
To highlight our sampling strategy's benefits for large-scale federated streaming data, we evaluate three key aspects: \textbf{numerical stability}, \textbf{wall-clock runtime}, and \textbf{sampling quality} in the following experiments, mimicking real-world scenarios on resource-constrained FL clients.
%
%

\begin{figure}[!htb]
    \centering
    \begin{minipage}[b]{0.48\textwidth}
        \centering        \includegraphics[width=0.94\linewidth]{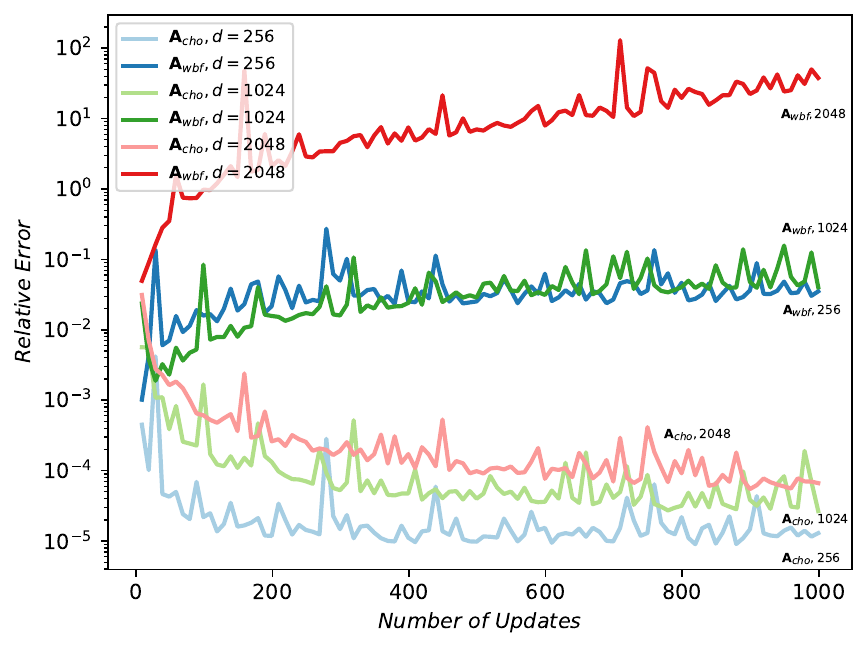}\vspace{-.5cm}
        \caption{Relative matrix reconstruction error comparison.}
        \label{fig:exp_1}
    \end{minipage}
    \hfill
    \begin{minipage}[b]{0.48\textwidth}
        \centering
        \scalebox{0.7}{
        \begin{tabular}{c|ccccccc}
        \toprule
        Method & Complexity & $d$ & \O \ Time\\
        &&& sec $\times 10^{-3}$ \\ 
        \hline
        & & 256 & $0.76 \pm 0.53$ \\
        Direct &  $\mathcal{O}(d^3)$ & 1024 & $16.11 \pm 3.65$ \\
        & & 2048 & $111.86 \pm 14.36$ \\ \hline
        & & 256 & $0.60 \pm 0.50$ \\ 
        Woodbury & $\mathcal{O}(r^2d)$ & 1024 & $10.96 \pm 2.52$ \\
        & & 2048 & $86.55 \pm 13.63$ \\ \hline
        & & 256 & $\textbf{0.60} \bm{\pm} \textbf{0.49}$ \\ 
        Ours & $\mathcal{O}(\frac{1}{3} d^3)$ & 1024 &$\textbf{6.68} \pm \textbf{1.54}$ \\
        & & 2048 & $\textbf{45.05} \pm \textbf{6.47}$ \\
        \bottomrule
        \end{tabular}
        }
        \caption{Algorithmic complexity and average wall-clock runtime comparison. Results calculated over three runs.} 
        \label{fig:exp_2} 
    \end{minipage}
\end{figure}

The \emph{first experiment} examines the low-rank update mechanism that iteratively recomputes the inverse tracking covariance matrix $\hat{\mathbf{\Sigma}}^{-1}$ after adding a new sample to $B$ in regards to its numerical stability over a large number of update cycles.
We therefore measure the reconstruction quality by means of relative error between the directly computed matrix $\mathbf{A}_{dir}$, updated by a randomized vector $\mathbf{v} \in \mathbb{R}^{d}$ such that $\mathbf{A}_{dir} = (\mathbf{A} + \mathbf{v}\mathbf{v}^T)^{-1}$,
and the corresponding matrices calculated employing low-rank update strategies based on both Woodbury formula $\mathbf{A}_{wbf}$ and Cholesky decomposition $\mathbf{A}_{cho}$ (ours). The relative error is subsequently defined as $\norm{\mathbf{A}_{wbf} - \mathbf{A}_{dir}}_F / \norm{\mathbf{A}_{dir}}_F$ and $\norm{\mathbf{A}_{cho} - \mathbf{A}_{dir}}_F / \norm{\mathbf{A}_{dir}}_F$, with $\norm{\cdot}_F$ employing the Frobenius norm.
The results of this experiment, traced over 1000 iterative updates with input dimensions $d \in \{256, 1024, 2048\}$, are shown in Fig.~\ref{fig:exp_1}.
The experiment demonstrates that our approach achieves faster convergence to a stable error and maintains consistently low relative reconstruction error rates throughout the iterative process across all input dimensions, in comparison to the Woodbury update-based method.
It is important to note that, unlike our technique, the Woodbury update-based method becomes unstable at higher input dimensions (see fluctuation for $d$ = 2048), underlining the future-proof suitability of our method for ever-increasing neural network models. 

In the \emph{second experiment} we evaluate our approach in terms of practical wall clock computation times on a CPU-based system.
Simulating a low-resource edge device without access to GPU acceleration, we follow exactly the same setup as explored in the first experiment.
The computation times are averaged for each dimension over the number of update cycles, see Fig.~\ref{fig:exp_2}.
\begin{wrapfigure}{r}{0.45\textwidth} 
  \centering\vspace{-0.35cm}
  \includegraphics[width=0.38\textwidth]{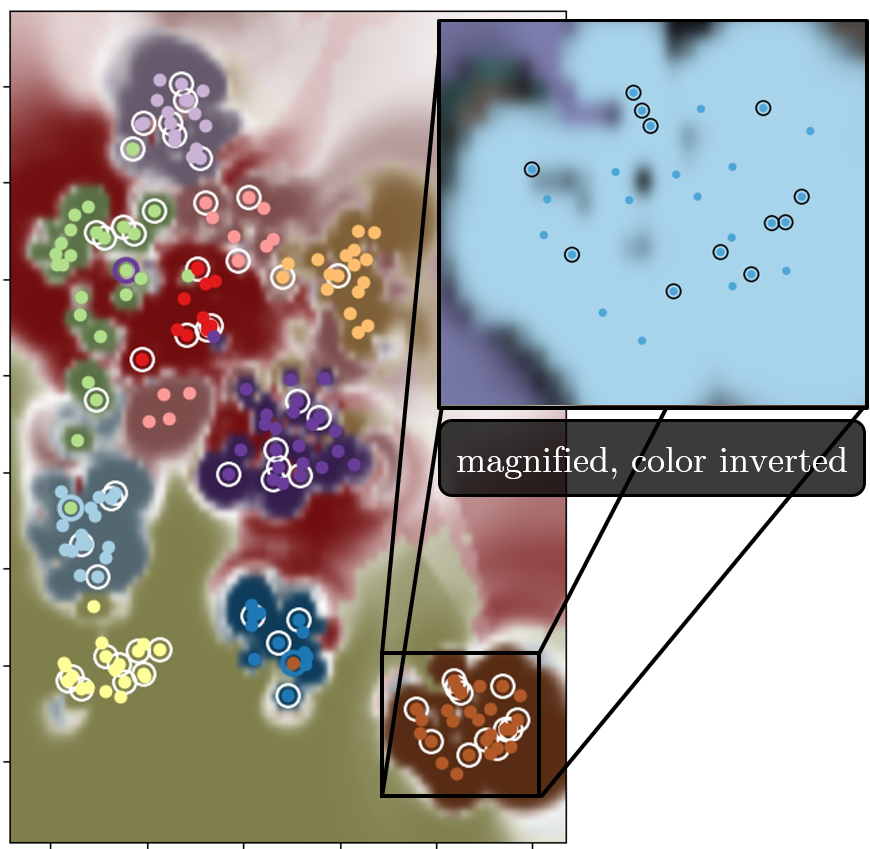}\vspace{-.3cm} 
  \caption{Model decision regions.}
  \label{fig:exp_3}\vspace{-0.35cm}
\end{wrapfigure}
\noindent
Although Woodbury updates (rank $r$=1) have lower theoretical operation counts than Cholesky-based low-rank techniques, they do not consider the nature of the calculations, granting our method a significant advantage in practical implementations:
Our approach operates exclusively on lower triangular forms, using cheap back-substitutions and avoiding complex operations for zero elements of the matrix.
This results in compute times that are on-par or faster in all scenarios.

The \emph{third experiment} visually confirms that our approach produces high-quality samples that can significantly contribute to model training optimization.
In this setup, the (client) model is first fine-tuned on CIFAR-10 training data as described in~\cite{roeder2024}. The stream sampling is subsequently performed with our see-only-once-strategy over the held-back test set of 10.000 images, cast as non-I.I.D. data stream under artificial feature drift as outlined in~\cite{saran2023}, with a labeling budget $k$ = 60.
We depict the model's decision regions and our selected data points (highlighted white circles) with DeepView~\cite{schulz2020} after adding randomly chosen samples for better illustration in Fig.~\ref{fig:exp_3}.
DeepView visualizes neural networks' decision functions in 2D by applying discriminative dimensionality reduction, thus revealing how our model classifies data. 
DeepView's reliability stems from its focus on features crucial for decision-making, offering precise insights into a model's behavior.
The plot confirms on the one hand that the sampling mass is well-distributed across the data stream by delivering a class-diversified batch $B$ and on the other hand that the decision-making algorithm tends to choose samples living near the edge of the model’s decision boundaries (see color-inverted, magnified area), providing relevant information for the training process.

\section{Conclusion}\label{sec_concl}
In this work, we introduced a novel sampling strategy for federated streaming data, driven by a decision making engine that iteratively performs low-rank updates on the Cholesky factor, efficiently recalculating the inverse covariance matrix which is employed to evaluate model training importance of an observation instantly.
The experimental results demonstrate that our method not only ensures numerical stability by reducing the accumulation of round-off errors over a large number of sampling iterations, but also minimizes computational overhead and enhances real-time system performance while selecting high-quality data points to support and improve model adaptation on federated clients.

\noindent
\emph{Future work} might explore methods to leverage the expert-labeled samples not only in the scope of local client personalization, but also for global FL model enhancements.
\emph{Code available at} \href{https://github.com/cairo-thws/fed_streaming}{github.com/cairo-thws/fed\_streaming}.


\begin{footnotesize}
\bibliographystyle{unsrt}
\bibliography{main.bib}

\begin{thebibliography}{10}

\bibitem{mcmahan2017}
B.~McMahan et~al.
\newblock Communication-efficient learning of deep networks from decentralized data.
\newblock In {\em Proc. 20th AISTAT}, volume~54, pages 1273--1282, April 2017.

\bibitem{rieke2020}
N.~Rieke.
\newblock The future of digital health with federated learn.
\newblock {\em Digital Medicine}, 3, 12 2020.

\bibitem{long2020}
G.~Long et~al.
\newblock Federated learning for open banking.
\newblock In {\em Fed. Learn. - Privacy and Incentive}, volume 12500 of {\em LNCS}, pages 240--254. Springer, 2020.

\bibitem{zhang2021}
H.~Zhang et~al.
\newblock End-to-end federated learning for autonomous driving vehicles.
\newblock In {\em 2021 IJCNN}, pages 1--8, 2021.

\bibitem{heusinger2023}
M.~Heusinger.
\newblock {\em Learning with high dimensional data and preprocessing in non-stationary environments}.
\newblock PhD thesis, Bielefeld University, Germany, 2023.

\bibitem{marfoq2023}
O.~Marfoq et~al.
\newblock Federated learning for data streams.
\newblock In {\em Proc. 26th AISTAT}, volume 206, pages 8889--8924. PMLR, 25--27 Apr 2023.

\bibitem{settles2009}
B.~Settles.
\newblock Active learning literature survey.
\newblock CS Tech. Rep. 1648, WISC, 2009.

\bibitem{saran2023}
A.~Saran et~al.
\newblock Streaming active learning with deep neural networks.
\newblock In {\em Proc. 40th ICML}, ICML'23. JMLR.org, 2023.

\bibitem{max1950}
M.~Woodbury.
\newblock Inverting modified matrices.
\newblock In {\em Mem. Rept. 42}, page~4. Princeton, 1950.

\bibitem{seeger2004}
M.~Seeger.
\newblock Low rank updates for the cholesky decomposition.
\newblock 2004.

\bibitem{roeder2024}
M.~Röder et~al.
\newblock {Crossing Domain Borders with Federated Few-Shot Adaptation}.
\newblock In {\em Proc. 13th ICPRAM}, pages 511--521, 2024.

\bibitem{golub2013}
G.~Golub et~al.
\newblock {\em Matrix Computations}, pages 338--341.
\newblock JHU Press, 20134.

\bibitem{schulz2020}
A.~Schulz et~al.
\newblock Deepview: Visualizing classification boundaries of deep neural networks as scatter plots using discriminative dimensionality reduction.
\newblock In {\em Proc., {IJCAI} 2020}, pages 2305--2311. ijcai.org, 2020.

\end{thebibliography}
\end{footnotesize}


\end{document}